\RequirePackage{colortbl}
\documentclass[12pt,lot,lof]{puthesis_undergraduate}

\usepackage{xcolor}
\usepackage{amsfonts}
\usepackage{amssymb}
\usepackage{amsmath}
\usepackage{amsthm}
\usepackage{latexsym}

\usepackage{tcolorbox}
\usepackage{graphicx}
\usepackage{url}

\usepackage{booktabs}

\newcommand{\bes}{\begin{equation*}}

\title{Lost in the Logic: An Evaluation of Large Language Models’ Reasoning Capabilities on LSAT Logic Games}
\submitted{April 2024}  
\author{Saumya Malik}
\advisor{Professor Danqi Chen} %

\abstract{
In this thesis, I evaluate the performance of Large Language Models (LLMs) on the Law School Admissions Test (LSAT), specifically the Logic Games section of the test. I focus on this section because it presents a complex logical reasoning task and thus is a valuable source of data for evaluating how modern, increasingly capable LLMs can handle hard logical reasoning tasks. I construct a dataset of LSAT logic games and their associated metadata, and extensively evaluate LLMs' performance in a Chain-of-Thought \cite{wei2023chainofthought} prompting setting. Given the weak performance in this setting, I explore other prompting frameworks on a smaller subset of the dataset, adapting ideas from Reflexion \cite{reflexion} to this task. This results in a substantially improved accuracy of 70\% for GPT-4 and 46\% for GPT-3.5 on this data subset, highlighting the capacity of LLMs to revise their logical errors, despite initially weak performance. Finally, I analyze the types of logic games that models perform better or worse on, as well as the types of logical errors I observe from human annotation, providing detailed insights on the logical reasoning capabilities of LLMs.

}

\acknowledgements{
Thank you to my advisor, Professor Danqi Chen, for your support and patience over the past two years, and for your thoughtful advising throughout this thesis. I owe my interest and future in research to you. 
}

\begin{document}
\chapter{Introduction}\label{ch:intro}  

In the wake of the release of GPT-4 \cite{openai2024gpt4}, Llama 2 \cite{touvron2023llama}, Claude 2 \cite{claude}, Mistral \cite{jiang2023mistral}, and other high-performance large language models (LLMs) all within the past year, there has been renewed interest in developing good evaluation benchmarks for LLMs. Older evaluation benchmarks may either be trivial or no longer usable, due to both saturation, dataset contamination, and overall simplicity. One important area for LLM evaluation is their reasoning capabilities, with much research aiming to uncover the extent to which LLMs can perform complex reasoning as opposed to merely reciting facts or complex explanations they have been trained on. To answer these questions, it is increasingly important to develop good evaluation benchmarks.

In their 2023 report on GPT-4, OpenAI reports an accuracy of 67\% on the Law School Admissions Test (LSAT), an impressively high number for a difficult exam that most aspiring law students spend months preparing for. OpenAI does not release further details or analysis of how, specifically, they prompted GPT-4 other than mentioning a multi-shot setting \cite{openai2024gpt4}. This motivates taking a closer look at GPT-4's performance on the LSAT, along with other LLMs. The Logic Games section of the LSAT, in particular, is an underutilized resource for qualitatively and quantitatively evaluating the performance of LLMs on tasks that require careful and complex logical deductions. 

In this thesis, I construct a dataset of LSAT logic games and probe several closed-source and open-source LLMs with different prompting strategies to evaluate how they perform on highly complex logical reasoning questions. I also qualitatively evaluate their performance along a number of analytical dimensions, deriving insights about particular types of logical errors LLMs are prone to. In particular, the contributions are fourfold:
\begin{enumerate}
\item The construction of a novel dataset containing all publicly-available LSAT logic games and metadata including their difficulty and game type. I release this dataset publicly on HuggingFace at this link \footnote{https://huggingface.co/datasets/saumyamalik/lsat\_logic\_games-analytical\_reasoning} to encourage further research.
\item An in-depth exploration of the limits of traditional Chain-of-Thought \cite{wei2023chainofthought} prompting for task of LSAT logic games, as well the potential applicability of Chain-of-Thought prompting to a particular subset of our dataset.  
\item An application and implementation of the Reflexion \cite{reflexion} framework to our LSAT logic games task, resulting in substantially improved accuracy compared to Chain-of-Thought prompting.
\item A quantitative and qualitative analysis of the \textit{types} of logic LLMs particularly excel at and fall short in.
\end{enumerate}

I find that Multi-Shot Chain-of-Thought prompting results in low overall accuracy on the LSAT logic games task, with GPT-4 achieving the best accuracy of only 33\% among the models I evaluate, with different models performing better or worse on different categorical types of logic games. This showcases the value of our dataset, as it presents a genuinely difficult logical task for LLM evaluation, but also highlights the limit of Chain-of-Thought prompting for exploring the limits of LLM capabilities on this task. 

Applying the framework from Reflexion \cite{reflexion}, though, which allows models the opportunity to reflect on their prior logical errors and reattempt a problem, results in a substantially improved accuracy of 70\% for GPT-4 and nearly 50\% for GPT-3.5 when taking the revised answer into account. This highlights both the impressive ability of LLMs to reflect and identify their logical errors, when given the chance, and the utility of such frameworks for evaluation as LLMs and evaluation metrics themselves become more complex.

In Chapter \ref{ch:background}, I provide background on the LSAT logic games section and prompt engineering as a field and discuss related works on work involving LLMs and the LSAT. In Chapter \ref{ch:implementation}, I detail the dataset construction and describe the various parameters a logic game question can have before discussing implementation of Chain-of-Thought prompting and my adaptation of Reflexion (which I call ``Self-Reflection"). Chapter \ref{ch:full_dataset} discusses and analyzes results of all LLMs I study over the full dataset. Chapter \ref{ch:ino} then discusses the Chain-of-Thought and Self-Reflection experiments I conduct on a subset of the full dataset that contains all logic games of the ``In-and-Out" variety. Chapter \ref{ch:ino} also analyzes and discusses specific logical errors GPT-3.5 and GPT-4 make in their responses to logic game questions, drawing insights on the kinds of logic these models struggle with. 

\chapter{Background and Related Work}\label{ch:background}  
\section{LSAT Logic Games}
\label{background:logic games}
The Law School Admission Test, or LSAT, is a standardized test that has been run by the Law School Admission Council (LSAC) since 1948. The current form of the LSAT consists of three types of sections: reading comprehension, logical reasoning (which is ``designed to evaluate your ability to examine, analyze, and critically evaluate arguments as they occur in ordinary language"), and analytical reasoning, more commonly referred to as logic games. 

Logic games are ``designed to assess your ability to consider a group of facts and rules, and, given those facts and rules, determine what could or must be true" \cite{lsac-lawhub}. Logic games vary widely in their content matter, type of logical question being asked, and difficulty, giving them several parameters worth analyzing. Beginning in August 2024, though, the logic games section of the LSAT will be removed. Nonetheless, the existing bank of logic games provide a great resource for evaluating the ability of LLMs to understand complex logical scenarios and answer questions that require simultaneously satisfying several logical constraints. I describe logic games in further depth and provide examples in Section \ref{sec:dataset} and in Appendix \ref{ch:sample Questions}.

\section{Prompting}
While foundational Large Language Models are incredibly powerful, how to craft questions (or ``prompts'') for them in a way that accesses their full power is still an active area of research known as prompt engineering. Prompt engineering is ``a relatively new discipline for developing and optimizing prompts to efficiently use language models (LMs) for a wide variety of applications and research topics" \cite{Saravia_Prompt_Engineering_Guide_2022}. Contemporary advances in prompt engineering thus provide relevant background and inform the approach of my thesis. One particularly relevant prompting strategy that I explore in this thesis is Chain-of-Thought Prompting \cite{wei2023chainofthought}.

\subsection{Chain-of-Thought Prompting}
\label{background:chain of thought}
Chain-of-Thought Prompting \cite{wei2023chainofthought} refers to explicitly telling a model to go step-by-step and explain its logic. It has been shown to be very helpful on more difficult question answer tasks that require step-by-step reasoning. In ``Multi-Shot" Chain-of-Thought prompting, we can prepend multiple examples of questions and accompanying step-by-step explanations to the prompt. These examples serve as demonstrations to further prompt the model to thoughtfully lay out its reasoning. In ``Zero-Shot" Chain-of-Thought, there are zero provided examples, and we can simply append ``Let's think step by step" to the prompt \cite{kojima2023large}. This has been shown to greatly improve model performance.

\section{Related Work}
\label{sec:related work}
\subsection{Logical Reasoning Datasets}
There are a number of existing logical reasoning datasets for evaluation, including LogiQA \cite{logiqa}, FOLIO \cite{folio}, LogicNLI \cite{logicnli}, and more. For the most part, these datasets evaluate the execution of simple logical rules. LSAT logic games have harder questions, more intricate multi-step logic that requires simultaneously satisfying many constraints at once. As such, they provide a valuable additional benchmark for seeing how increasingly capable LLMs perform on difficult multi-step logic questions. LSAT logic games thus provide a task that allow us to to evaluate how LLMs execute multi-step logic with many constraints, but also allows us to see at a micro-level whether models are able to execute even simple logical rules when presented in a difficult context. 

\subsection{LSAT and Language Models}
A few prior works look at the LSAT. Zhong et al. collect and release a dataset of LSAT logic game questions (that is used by most research on the LSAT), but their approach (creating a non-LLM system for solving LSAT questions) is different from my motivation of evaluating how LLMs perform on the LSAT \cite{zhong2021arlsat}\cite{wang2022lsat}. Furthermore, their dataset does not include valuable metadata (Problem/Game Difficulty, Problem/Game Type) that guides much of my approach and my analysis. I hope releasing my dataset will inspire future work that makes use of this rich metadata.

Additionally, Ye et al. \cite{SatLM} look at improving the performance of LLMs on this aforementioned LSAT dataset, but their approach focuses on using LLMs to formalize the problem, which then gets solved by an external solver. My approach, though, focuses on probing different prompting methods to see how the LLMs themselves do on solving the puzzles. Furthermore, while Ye et al. report a baseline of 23\% accuracy by GPT-3.5 on the LSAT logic games, they do not analyze this result over more prompting settings. My work both provides analysis to elucidate what might be giving rise to the 23\% number and also probes alternate prompting frameworks that result in LLM performance that improves on their reported baseline.


\chapter{Dataset Construction and Implementation}\label{ch:implementation}
\section{Dataset Construction}
\label{sec:dataset}
The first major contribution of my project is the construction of a dataset of all LSAT Logic Games that are available for free online. I release this dataset to encourage future research at: https://huggingface.co/datasets/saumyamalik/lsat\_logic\_games-analytical\_reasoning. To create my dataset, I collected every logic game question from every past LSAT test that has been released for free publicly (many tests are behind a paywall and thus would not be releasable as part of my dataset). There are 18 tests accessible on the web with free accounts—five on the official LSAC LawHub Website \cite{lsac-lawhub}, and thirteen on Khan Academy \cite{khan-academy}. Each test has four game setups, each with 5-6 questions that follow, for a total of 23 questions per test (with the exception of one retracted question in one past exam). In total, this amounts to 413 questions. From the LSAC website and Khan Academy, I copied over each question's setup rules, question and answer choices, and correct answer. I also curated explanations for twenty of the questions, adapting explanations from LSATHacks \cite{lsat-hacks} and Khan Academy. These explanations serve as demonstrations for Chain-of-Thought prompting.

Each question additionally has several parameters relating to the type and difficulty of the logic game as a whole and the question itself. These are not official parameters and categories that LSAC tags the problems with, but rather categories that well-established and well-respected LSAT prep sites online tag problems with. As such, they can be used as ground-truth human annotations for the dataset. For each question, I found values for each parameter from the online blogs 7SagePrep \cite{7sage} and PowerScore \cite{powerscore}. This is a novel contribution— no other LSAT dataset compiles these annotations, which provide valuable dimensions for prompting and analysis.
I will now describe each of these parameters below along with some examples, and additional examples can be found in Appendix \ref{ch:sample Questions}.
\subsection{Logic Game Difficulty and Problem Difficulty}
Logic Games are rated for difficulty on a scale of 1 (easiest) to 5 (hardest). Within each game, each problem is similarly rated for difficulty on a scale of 1 to 5. While often there is correlation between Game Difficulty and Problem Difficulty—it is common for most problems in a Game of Difficulty 5 to also have very high difficulty—it is possible for Games of Difficulty 5 to have Problems of Difficulty 2 and vice versa. These labels are taken from 7SagePrep \cite{7sage}. 

\subsection{Logic Game Type}
\label{sec:logic game type}
Logic Games can come in many types. Each type may differ in the nature of assignment they are asking for and in the nature of their constraints. Logic Game types can be broadly grouped into In-and-Out games, Sequence games, and Grouping games.\footnote{Except when otherwise indicated, example questions are from Khan Academy \cite{khan-academy}.}
\pagebreak
\subsubsection{In-and-Out Games} In-and-Out games involve binary decisions— each entity can take on one of two values (``in'' or ``out") for a particular trait. These games have lots of \textit{If/Then} conditionals (that can be formalized as implications) in their constraints and questions. An In-and-Out question may involve assigning cookbooks to be published in either fall (which I can set to be ``in") or spring (``out"):
\newtcolorbox{promptbox}[1]{
        boxrule = 1pt,
        fontupper = \small\fontseries{l}\tt,
       fonttitle = \bf\color{black},
       arc = 5pt,
        rounded corners,
        colframe = black,
        colbacktitle = white!90!blue,
        colback = white!97!blue,
        title = #1,
}

\begin{minipage}[t]{\textwidth}
    \vspace{0pt} 
    \begin{singlespace}
    \begin{promptbox}{In-and-Out Question}
A publisher is planning to publish six cookbooks-K, L, M, N, O, and P-over the course of the next year. Each cookbook will be published in one of two seasons-fall or spring-subject to the following conditions:\\\\
-M and P cannot be published in the same season as each other.\\
-K and N must be published in the same season as each other.\\
-If K is published in the fall, O must also be published in the fall.\\
-If M is published in the fall, N must be published in the spring.
\\\\
If M is published in the fall, which one of the following is a pair of cookbooks that could both be published in the fall along with M?\\
(A) K and O\\
(B) L and N\\
(C) L and O (correct)\\
(D) N and P\\
(E) O and P
    \vspace{0.445cm}
    \end{promptbox}%
    \end{singlespace}
\end{minipage}%
\pagebreak
\subsubsection{Sequence Games} Sequence games involve coming up with a logically valid sequence of objects given constraints on their ordering. More formally, they have many Partial Order Relations \cite{davey2002introduction} as constraints. For example, a sequence game could involve ordering six different layers of cake according to rules specifying the relative ordering between layers as well as rules about their absolute ordering:

\begin{minipage}[t]{\textwidth}
    \vspace{0pt} 
    \begin{singlespace}
    \begin{promptbox}{Sequence Question}
   A cake has exactly six layers-lemon, marzipan, orange, raspberry, strawberry, and vanilla. There is exactly one bottom layer (the first layer), and each succeeding layer (from second through sixth) completely covers the layer beneath it. The following conditions must apply:\\\\
-The raspberry layer is neither immediately above nor immediately below the strawberry layer.\\
-The marzipan layer is immediately above the lemon layer.\\
-The orange layer is above the marzipan layer but below the strawberry layer.
\\\\
Which one of the following could be an accurate list of the layers of the cake, from bottom to top?\\
(A) lemon, marzipan, orange, strawberry, vanilla, raspberry    (correct)\\
(B) lemon, marzipan, orange, strawberry, raspberry, vanilla\\
(C) marzipan, lemon, raspberry, vanilla, orange, strawberry\\
(D) raspberry, lemon, marzipan, vanilla, strawberry, orange\\
(E) raspberry, orange, lemon, marzipan, strawberry, vanilla
    \vspace{0.445cm}
    \end{promptbox}%
    \end{singlespace}
\end{minipage}%
\pagebreak
\subsubsection{Grouping Games} Grouping games involve grouping entities by coming up with a logically valid assignment of entities to groups, given constraints. For example, this could look like assigning six employees to three booths:

\begin{minipage}[t]{\textwidth}
    \vspace{0pt} 
    \begin{singlespace}
    \begin{promptbox}{Grouping Question}
For an antiques fair at the local civic center, the fair's manager must assign each of six employees-Frank, Gladys, Hal, Keisha, Laura, and Mike-to one of three information booths-the organizers booth, the retailers booth, and the visitors booth. Each booth must be assigned at least one employee. The assignments are constrained as follows:\\\\
-The retailers booth must have more employees than the visitors booth.\\
-Neither Frank nor Keisha can be assigned to the visitors booth.\\
-Neither Gladys nor Hal can be assigned to the organizers booth.\\
-Gladys and Mike must work at the same booth as each other.\\\\
If Hal is assigned to a booth with exactly one other employee, then which one of the following could be true?\\
(A) Laura is assigned to the organizers booth.\\
(B) Frank is assigned to the retailers booth. (correct)\\
(C) Hal is assigned to the retailers booth.\\
(D) Laura is assigned to the retailers booth.\\
(E) Gladys is assigned to the visitors booth.
    \vspace{0.445cm}
    \end{promptbox}%
    \end{singlespace}
\end{minipage}%
\\\\

There are many hybrid and variant categories that combine aspects of each of these three broad categories. Hybrid categories include ``Group Sequence," ``In-and-Out Sequence," and more, with these labels also taken from 7SagePrep. Examples of each hybrid game can be found in Appendix \ref{ch:sample Questions}.
\subsection{Problem Type}
There are many sub-dimensions of Problem Type, and one problem may take on values for each sub-dimension:
\begin{itemize}
\item \textbf{Locality:} A problem can be ``local" or ``global." ``Global" problems ask for logical deductions based on the global set of rules and setup of the problem (like the Sequence question above), while ``local" problems introduce a new local constraint and posit how this added constraint affects or narrows the possibility space (like the Grouping and In-and-Out questions above).
\item \textbf{Degree of Truth:} Many questions are of the form ``\textit{Which of the following must/could/cannot be true?"} I refer to this as a problem's degree of truth. (All three questions above are ``could be true" questions.) 

\item \textbf{Others:} There are several other problem types, none of which are mutually exclusive with each other, including rule substitution (\textit{Which of the following rules, if replaced with X constraint, achieves the same outcome?}), and ``except" questions (\textit{All of the following could/must be true \textbf{except}?}). These problem types do not feature heavily in my analysis.
\end{itemize}
These labels are taken from the PowerScore Forum \cite{powerscore}, but with the caveat that in my experience, the labels for Problem Type are occasionally incomplete (e.g., a problem that is both ``local" and ``could be true" might only be marked in PowerScore as ``local"). I correct these errors as I notice them, but some labels may still be incomplete.

\subsection{Answer Explanations}
\label{sec:ds_explanations}
One challenge with selecting explanations to serve as Chain-of-Thought demonstrations is that reasoning paths can vary widely by problem— for some problems, it is better to make deductions from the rules and problem statement before looking at any answer choices, whereas for other problems, it may be better to look at each answer choice and how it relates to the constraints one-by-one. This diversity of reasoning paths likely ends up being a limitation of Chain-of-Thought prompting, as I discuss when analyzing the results of extensive Chain-of-Thought experiments in Section \ref{sec:cot_weakness}. 

Nonetheless, crafting explanations to have a consistent high-level structure to their problem-solving is helpful. Khan Academy explanations, in particular, follow a consistent high-level structure of analyzing the problem setting (categorizing the logic game), making deductions based on constraints, then looking through each answer choice and making deductions along the way. I adapt these explanations slightly and reformat them to have this consistent formatting, with some added sign-posting to guide the model's output formatting. I always add a concluding ``\texttt{Therefore, the correct letter answer is: \{Correct Answer\}.}" to the end of the demonstration to facilitate effective answer extraction. I verify that the models emulate this answer formatting well. As an example, Figure \ref{fig:explanation} shows the explanation to the In-and-Out question above.

\begin{figure}
\include{Figures/ex_explanation}
\caption{Sample Explanation, Adapted from Khan Academy} \label{fig:explanation}
\end{figure}

I release this annotated dataset, along with the compiled explanations, to facilitate future research.

\section{Experimental Setup}
\label{sec:experimental setup}
\subsection{Models}
I sought  to evaluate how models would do if prompted with a logic game setup, question, and multiple letter answer choices. I prompted GPT-3.5 and GPT-4 via the OpenAI API.  For these experiments, I used \verb|gpt-3.5-turbo| (the January 25th version) and \verb|gpt-4-1106-preview|, which at the time of my experiments, was the most recent version available. I also prompted Claude 2 via the Anthropic API (but queried Claude 2 for only one prompting setting given pricing constraints). As for open source models, I prompted both Llama2-7b and Mistral-7b (significantly smaller models) using their versions on HuggingFace.
\subsection{Experimental Settings}
Since LLM bias towards particular answer labels has been documented for Multiple Choice Questions \cite{pride}, for each question, I ran 5 permutations, rotating the answer options such that the correct answer choice appears in each letter position once. This mitigates the effect of any positional bias in results. 

Except when otherwise indicated, I set sampling temperature to 0 to reduce variance, since my goal is to uncover the true inclination of models, and answers were highly inconsistent in initial experimentation with temperature greater than 0. I will now describe each prompting setting. 

\subsection{Prompting Settings - Full Dataset}
\label{sec:imp_promptsettings}
\subsubsection{Zero-Shot Chain-of-Thought Prompting}
Kojima et al. showed that even Zero-Shot Chain-of-Thought prompting can improve model performance over standard Zero-Shot prompting, simply by explicitly prompting the model to reason step by step \cite{kojima2023large}. I implemented Kojima et al.'s two-step Zero-Shot Chain-of-Thought formulation. The first step, ``Reasoning Extraction," prompts the model with the question followed by ``\texttt{Let's think step by step}.". Then the second step, ``Answer Extraction," appends the response the model generated in the first step to the original prompt, along with the final formatting prompt, ``\texttt{Therefore, the correct letter answer is:}". This procedure encourages the model to carefully explain its reasoning.
\subsubsection{Multi-Shot Chain-of-Thought Prompting}
I also conducted Multi-Shot Chain-of-Thought Prompting \cite{wei2023chainofthought}, which prepends the question with demonstrations of questions and step-by-step explanations. I adapted the explanations from Khan Academy, LSATHacks, and PowerScore, as described in Section \ref{sec:ds_explanations} above. In particular, since enumerating over all possible examples for the whole dataset is infeasible and too ad hoc in its approach, I first look at smaller subsets of the datasets to find the example demonstrations that perform the best, and sample from this refined set for prompting over the whole dataset. In the next subsection, I describe the set of experiments I perform on a particular subset of the dataset— the subset containing all 43 ``In-and-Out" questions (including hybrid In-and-Out questions).
\subsubsection{Summing Over Underlying Logits}
Open source models additionally give access to logits; I eliminate positional bias by evaluating their performance on questions by summing over the logits they produce for each answer choice evaluated independently, as \texttt{lm-evaluation-harness} does for certain Multiple Choice tasks \cite{eval-harness}. For a particular question, for each answer choice I do the following: I append the answer choice to the question statement and feed it into the model. I then sum over the logits of the answer choice tokens, thereby capturing the model's likelihood it assigns to that answer choice. I do this for all five answer choices independently, and select the answer choice with the highest likelihood to be the model's selected answer. This setting works far better for open source models than prompting with Multiple Choice questions, so I only report the results for this method of prompting. I do not implement this for closed-source models since they do not give access to full output logits. 

\subsection{Experiments on In-and-Out Subset}
\label{sec:ino_implementation}
Initially motivated by strong GPT-4 Zero-Shot Chain-of-Thought performance on In-and-Out questions in particular, I focused on the subset of In-and-Out Questions (including hybrid questions) from the full dataset to see if I could further improve performance on this subset with multi-shot Chain-of-Thought Prompting and other strategies. This subset amounts to 43 questions, approximately 10\% of the full dataset. In addition to systematically exploring multi-shot Chain-of-Thought prompting on this subset, I also adapt Reflexion \cite{reflexion} to this problem setting and conduct a thorough analysis of logical errors.

\subsubsection{Multi-Shot Chain-of-Thought}
For my multi-shot Chain-of-Thought experiments, I created a pool of 5 In-and-Out Demonstrations, carefully hand picking problems to be not too easy\footnote{Preliminary exploration discovered that Easy demonstrations elicited terse incorrect logic from the model.} and to span a variety of Hybrid game types. (Since they are used as demonstrations, these five questions are removed from the In-and-Out dataset.)

I then conducted k-shot prompting for k = 1 to 5, at each level taking $k$ prompts from my pool of demonstrations. For each $k$, I took 5-9 combinations of $k$ demonstrations from the overall pool. Querying GPT-3.5\footnote{I conducted these numerous exploratory experiments on GPT-3.5 to save costs, assuming the relative strength of prompts would generalize to GPT-4.} with many prompts (for each $k$) is crucial for finding a prompt that works effectively.

\subsubsection{Self-Reflection}
\begin{figure}
\centering
\includegraphics[width=\textwidth]{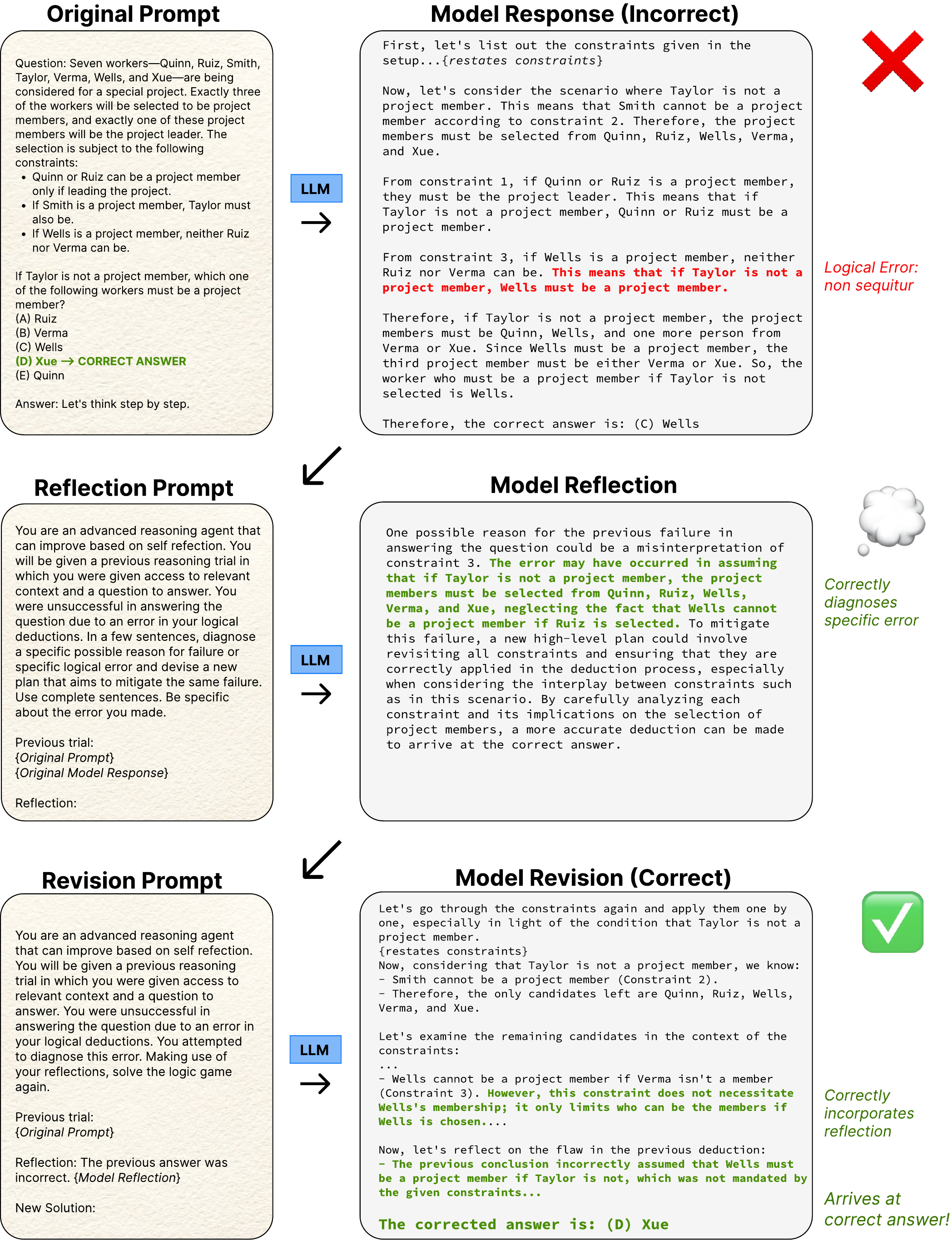}
\caption{Walkthrough of a Successful GPT-4 Self-Reflection}
\label{fig:reflection}
\end{figure}

In their paper ``Reflexion: Language Agents with Verbal Reinforcement Learning," Shinn et al. propose ``a novel framework to reinforce language agents not by updating weights, but instead through linguistic feedback" \cite{reflexion}. I apply the framework and ideas behind Reflexion \cite{reflexion} to see whether GPT-3.5 and GPT-4 can improve their accuracy on the In-and-Out dataset if given linguistic feedback from their environment, where feedback takes the form of indicating to the model that its answer was incorrect (but not revealing the correct answer). In other words, I explore if models are capable of Self-Reflection— identifying their own errors and getting the question right on a second try. I call this implementation Self-Reflection to distinguish it from Reflexion since this implementation is not a direct implementation of Reflexion (this implementation keeps no memory buffer for future trials, as one difference), but rather is inspired by the Reflexion framework. In particular, my implementation has 3 steps:
\begin{enumerate}
    \item \textbf{Prompt}: Prompt the model with the question and get its response.
    \item \textbf{Reflect}: If the response is incorrect, provide ``environment feedback" that the response is incorrect and ask it to reflect on what went wrong and make a plan to solve the question correctly next time.
    \item \textbf{Revise}: Provide the model its reflection and elicit a response.
\end{enumerate}
A successful example walking through the exact prompts for each of these three steps, along with GPT-4's initial response and successful reflection and revision, is illustrated in Figure \ref{fig:reflection}.

I made a few modifications to Reflexion's prompts to adapt it to our logic game task. For one, I changed the second step's Reflection prompt to ask for a \textit{specific} logical error (as opposed to a \textit{high-level} plan in the original implementation) after observing that asking for a high-level plan did not lead to successful revision on our task. I also chose to omit the model's original incorrect response in the revision prompt, only giving the model the Question and the Reflection in Step 3. I did this because I observed that providing the original incorrect response heavily biases the model toward producing that same incorrect response again.

For this experimental setting, I set temperature $>$ 0 to encourage the model to be creative and allow it to generate a new revised thought that differs from the original one it proposes. I implement Self-Reflection for only the smaller In-and-Out subset of the full dataset so that I can feasibly annotate each stage in the Self-Reflection procedure, making note of when reflections are correct and revised answers correctly implement reflections to arrive at a correct answer. For each question, I run Self-Reflection on one of the five question orderings. Future work can extend this promising prompting framework to the full dataset.

\subsubsection{Logical Error Framework}
Finally, I use our In-and-Out dataset to qualitatively evaluate the logical capabilities (and errors) of GPT-3.5 and GPT-4. I manually annotate a subset of 20 In-and-Out questions. For each question, I do the following:
\begin{enumerate}
    \item Read the constraints and draw logical conclusions from them (e.g, given an implication, observe that the contrapositive would also hold. Or, given two related implications, apply transitivity, if relevant.)
    \item Solve the problem by hand, tracing out the logical deductions as they come. Make note of what logical deductions lead to the elimination of each incorrect answer choice, thus establishing what logical \textit{errors} would be indicated by a model selecting that choice. (As a point of comparison for how difficult these problems are, these first two steps together take us, on average, five minutes per problem.) 
    \item Look carefully at GPT-3.5's solution to the problem, noting line by line both correct and incorrect logical deductions. For each deduction, categorize the logical rule involved.
    \item Repeat Step 3 for GPT-4.
\end{enumerate}

\chapter{Full Dataset Results and Analysis}\label{ch:full_dataset}
In this chapter, I present the models' accuracy averaged over the full 413 question dataset. I then dive into analyzing the performance of models along particular parameters of the dataset, making use of the metadata I compiled from the web in Section \ref{sec:dataset}. In this analysis, I observe trends of particular Logic Game Types and thus types of logic that models have a preference for, also drawing attention to differences between models. Finally, I conclude with an analysis of dataset contamination and why I am optimistic that any potential dataset contamination would not impact the results and insights from this thesis.

\section{Experimental Results: Summary}
In this section, I report the average accuracy of each model over the entire dataset. First, a few notes on implementation. Due to pricing constraints, I only query Claude 2 for the Zero-Shot Chain-of-Thought setting, which I select because it was the best-performing setting across other models. I verify over several trials that multi-shot Chain-of-Thought prompting does not improve performance on the \textit{whole} dataset for GPT-3.5, Mistral, and Llama, so I report accuracies only for the zero-shot and one-shot Chain-of-Thought settings. Later in \ref{ch:ino}, I explore multi-shot Chain-of-Thought prompting on a \textit{subset} of the data and find strong benefits, and I also discuss likely limitations of using multi-shot Chain-of-Thought prompting for the whole dataset. The results are summarized in Table \ref{tab:accuracy} below.

\begin{table}[htbp]
\caption{Average Accuracy on Dataset For All Models} \label{tab:accuracy}
\vspace{-10pt}
\begin{center}
\begin{tabular}{lllllll}
\hline \hline
Prompt &   &   GPT-3.5 & GPT-4 & Claude 2 & Mistral-7b & Llama2-7b \\
\hline Standard & 0-shot & 21.3 & 30.0 & n/a & 24.2 & 21.1\\
\hline Chain-of-Thought &  0-shot & 23.2 & 33.0 & 25.7 & 23.8 & 20.5 \\
& 1-shot & 23.1 & 28.3 & n/a & 23.7 & 21.1 \\
\hline \hline
\end{tabular}
\end{center}
\end{table}
These results are fascinating— for one thing, they confirm that LSAT logic games are a \textit{hard} task even for modern LLMs, with some models performing only slightly above random (corresponding to 20\% since there are five answer choices). However, GPT-4 still performs fairly well, achieving an accuracy of 33\% in the zero-shot Chain-of-Thought setting. Claude 2 achieves the next highest accuracy of 25.7\%. Despite being a much smaller model (only seven billion parameters), Mistral achieves 24.2\% accuracy in standard zero-shot prompting, not far behind Claude 2's performance. It is important to note that the results from the closed-source and open-source models are not directly comparable, since I evaluate the open-source models by summing over logits as described in Section \ref{sec:imp_promptsettings}. Given this, it is also unsurprising that open-source models did not perform better in the zero-shot Chain-of-Thought setting, which would amount to just appending ``\texttt{Let's think step by step.}" because I do not prompt the open-source models for generation.

Given that LSAT logic games are very hard overall, it would also be valuable to see how LLMs perform specifically on Easy problems, or problems with Problem Difficulty level 1. I report the models' average accuracy on the 65 Easy problems in our dataset in Table \ref{tab:easy_accuracy} below: 
\begin{table}[htbp]
\caption{Average Accuracy on Easy Problems For All Models} \label{tab:easy_accuracy}
\vspace{-10pt}
\begin{center}
\begin{tabular}{lllllll}
\hline \hline
Prompt &   &   GPT-3.5 & GPT-4 & Claude 2 & Mistral-7b & Llama2-7b \\
\hline Standard & 0-shot & 23.4 & 35.1 & n/a & 33.8 & 20\\
\hline Chain-of-Thought &  0-shot & 28.3 & 39.1 & 33.2 & 32.1 & 19.5 \\
& 1-shot & 27.6 & 33.3 & n/a & 32.2 & 19.8 \\
\hline \hline
\end{tabular}
\end{center}
\end{table}
We can see from this table that GPT-4 achieves a fairly high accuracy of 39.1\% on Easy questions. Interestingly, Llama 2 sees no improved performance on Easy questions, but this is unsurprising given its overall accuracy was already at random.

\section{Analysis by Parameter}
In this section, I analyze model performance by Problem Difficulty and Game Type. I also analyzed performance across \textit{Game} Difficulty and \textit{Problem} Type but observed no significant noteworthy trends.

\subsection{Problem Difficulty}
\begin{figure}
\caption{Accuracy by Problem Difficulty} \label{fig:prob_difficulty}
\vspace{0pt}
\centering
\includegraphics[width=\textwidth]{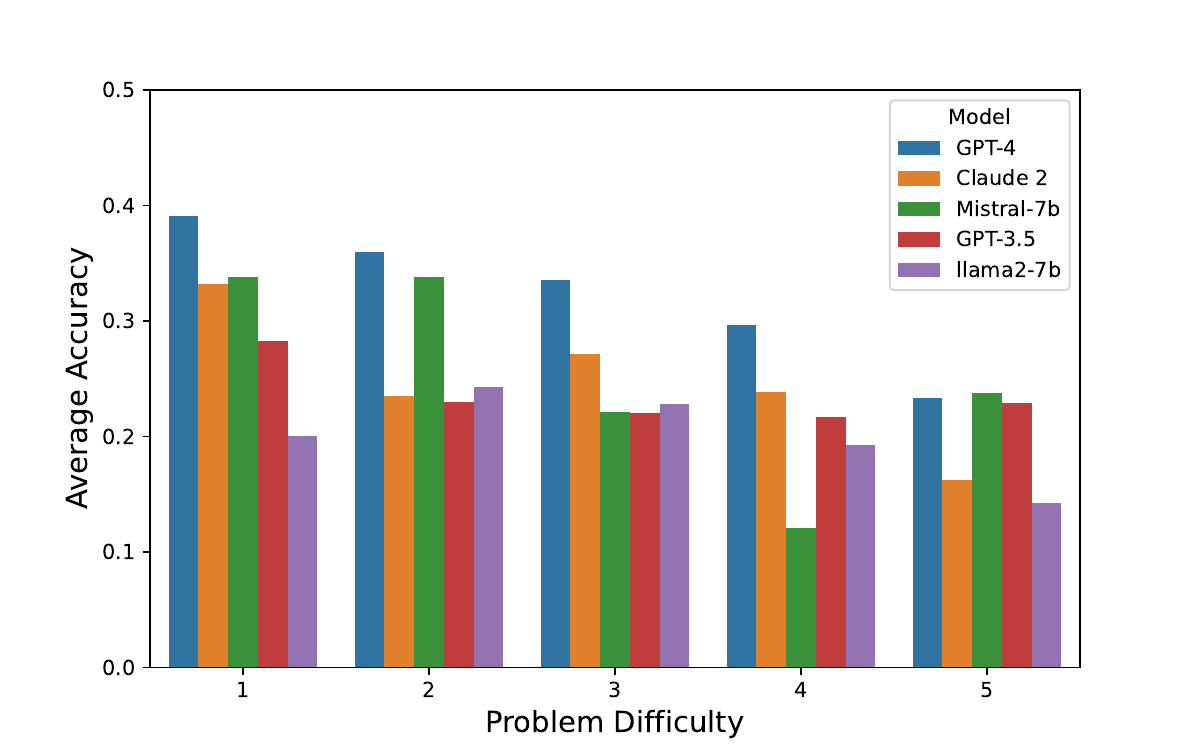}
\end{figure}

First, I analyze model performance by Problem Difficulty. As we can see in Figure \ref{fig:prob_difficulty}, all of the models, for the most part, have higher accuracy on easier problems and lower accuracy on higher problems, with this trend particularly salient for GPT-4 and Claude 2 (which are also the overall best-performing models). For reference, there are 64 problems of difficulty level 1, 74 of level 2, 149 of level 3, 83 of level 4, and 42 of level 5. This suggests that overall, at a high-level, humans and LLMs roughly find the same kinds of logic problems difficult or easy.

\subsection{Game Type}
\begin{figure}
\caption{Accuracy by Game Type} \label{fig:game_type}
\vspace{0pt}
\centering
\includegraphics[width=\textwidth]{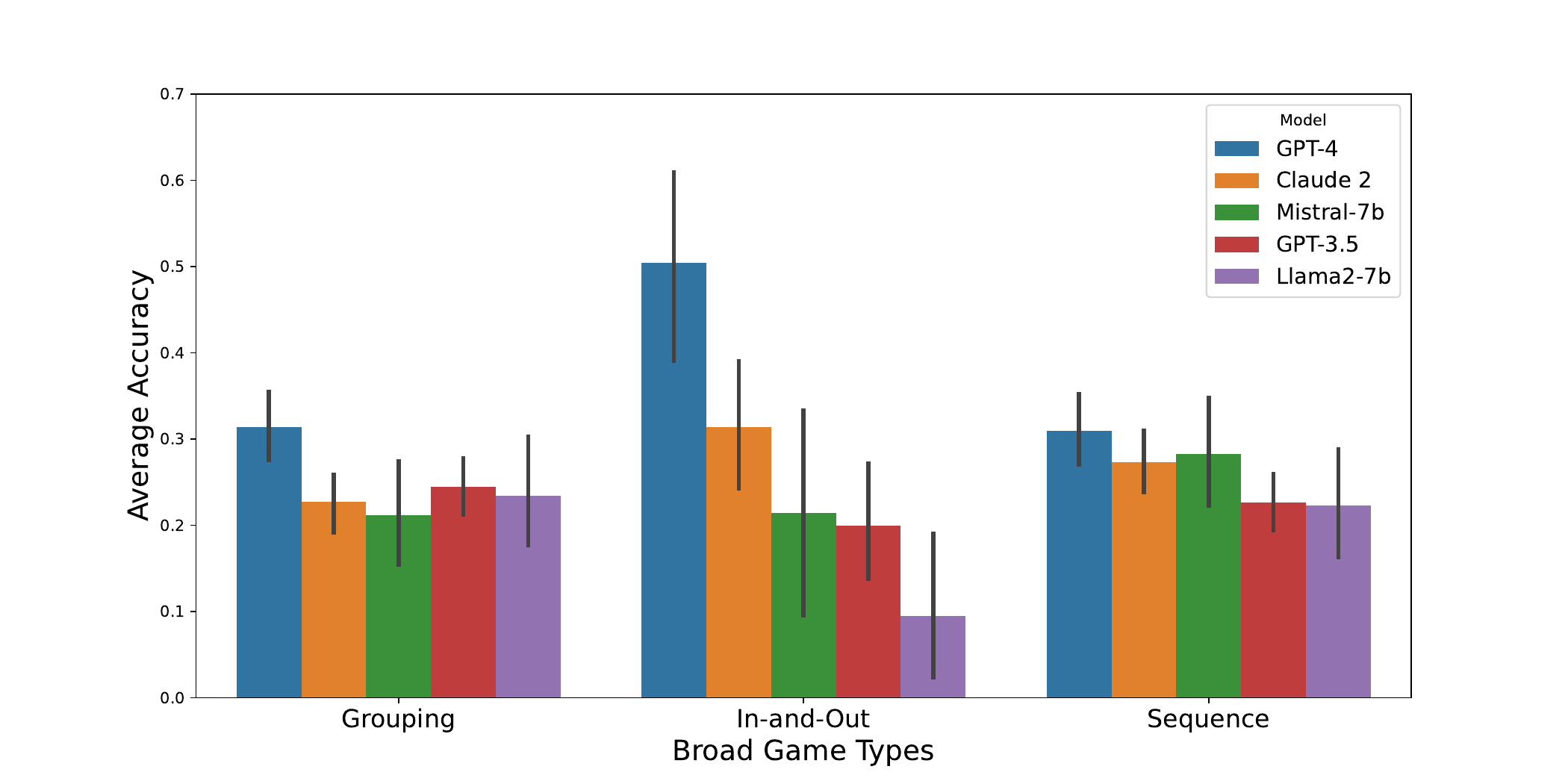}
\end{figure}

By looking at performance based on our broad categories for Game Type—Grouping, In-and-Out, and Sequence—in Figure \ref{fig:game_type}, we see that models perform significantly better on particular types of games than others. For reference, there are 181 Grouping game problems, 48 In-and-Out game problems, and 184 Sequence game problems.

In particular, GPT-4 performs \textit{significantly better} on In-and-Out games (which have lots of implications as constraints and require deft use of implication logic and transitivity to solve) than Grouping games and Sequence games (which require satisfying partial order relations). We can also see that GPT-4 significantly outperforms all other models on In-and-Out games and Grouping games. Furthermore, Claude 2 performs better on Sequence games than Grouping games, while GPT-4 shows no difference. The rest of the models do not seem to have a significant performance gap on one game over the other, except for Llama 2, which curiously performs significantly worse on In-and-Out games than the other two types. 

Of course, some of this variation in performance across different Game Types can be attributed to particular Game Types being easier in general in the dataset, potentially reflecting that they may be easy to humans. For reference, Grouping games have the highest average difficulty of 3.05/5, compared to 2.64/5 for In-and-Out and 2.78/5 for Sequence games. It is interesting to note from Figure \ref{fig:game_type} that models largely perform best on In-and-Out games, in accordance with human judgment, but that models other than Claude 2 do not exhibit significant preferences for Grouping games versus Sequence games despite the large difficulty gap between the two.

\begin{figure}
\caption{Accuracy by Game Type on Difficulty 3 Problems} \label{fig:game_type_normalized}
\vspace{0pt}
\centering
\includegraphics[width=\textwidth]{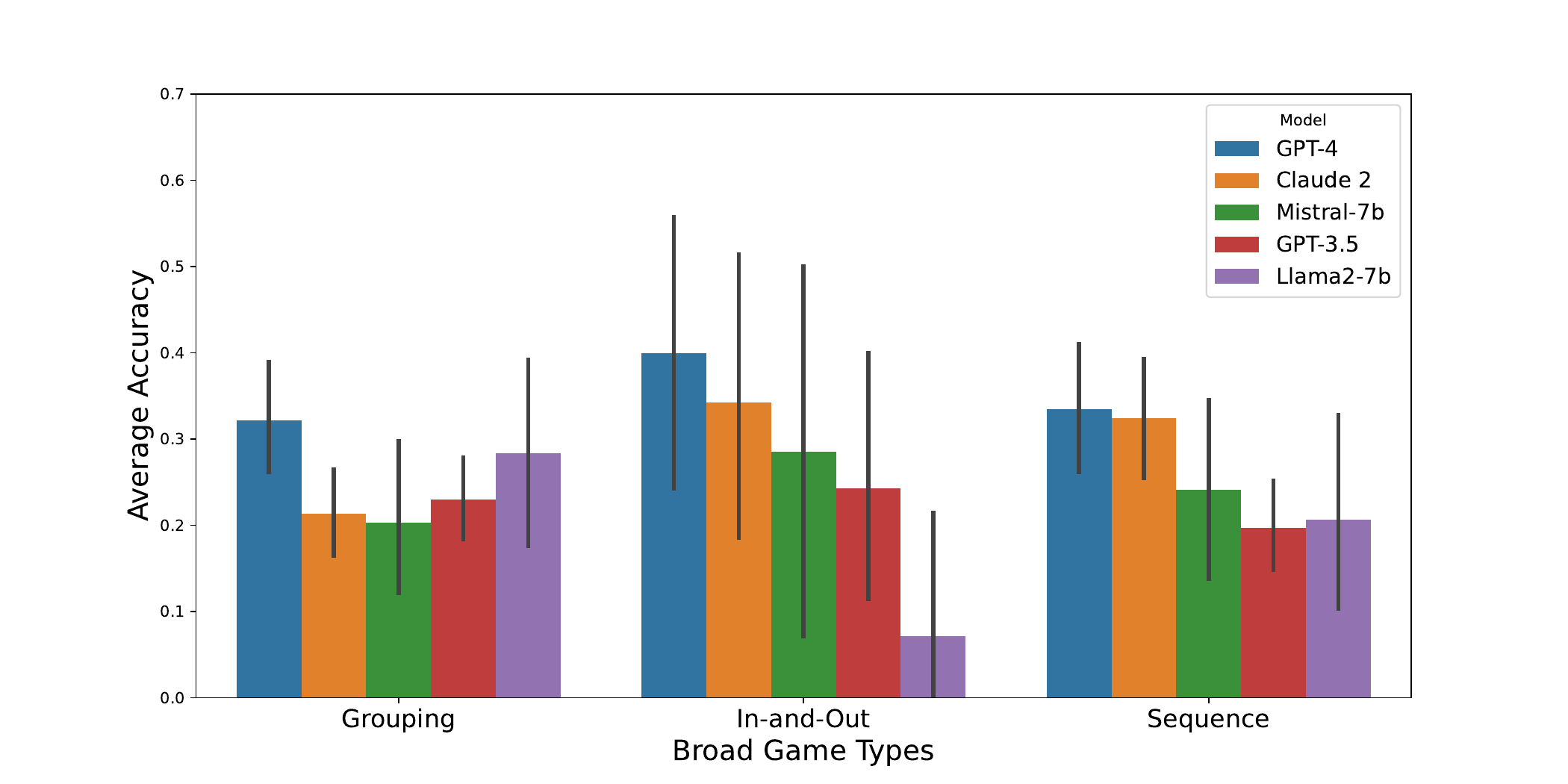}
\end{figure}

Additionally, then, to further tease out whether models have a preference for particular logical game types that goes beyond the ground truth difficulty of these games, I analyze accuracy by game type over all 149 problems of difficulty level 3 (selected as the difficulty level with the greatest number of problems) to normalize for problem difficulty. I display the results in Figure \ref{fig:game_type_normalized}.
It is interesting to note that even when normalized for problem difficulty, Claude 2 maintains its significantly higher performance on Sequence games over Grouping games! And while no longer statistically significant when normalized for difficulty, GPT-4 still has a higher performance on In-and-Out games. This provides valuable insight— Claude 2 may be better at ordering logic than grouping logic, while GPT-4 may be better at logically navigation implications (in In-and-Out games) than ordering or grouping logic. Motivated by GPT-4's strong performance on In-and-Out games, I explore this subset of the dataset in the next chapter. But first, I conclude this chapter by analyzing whether there are impacts from potential dataset contamination.

\section{A Note on Potential Dataset Contamination}
In this section, I analyze the results to detect whether there is an impact of dataset contamination on the LLMs' pretraining data. I follow the insights from Oren, Meister et al. \cite{oren2023proving}, who show that if the impact of dataset contamination is strong, models should have a preference toward data (or answer choices) being presented in their  ``canonical" orders as they appeared on the web. I conclude that the impacts of dataset contamination are not strong for the LSAT logic games task.

As with all data taken from the web, it is possible that some portion of our dataset might have been included in LLMs' training data. Even if some portion of our dataset is present in the training data, I argue that the results of this thesis still hold. In particular, I observe that any potential dataset contamination likely does not have significant impact on the models' performance on the task, which is fairly low to begin with. 

In their 2023 technical report for GPT-4, OpenAI evaluates GPT-4 on the LSAT and disclose that 39\% of their LSAT evaluation data was present in their training data \cite{openai2024gpt4}, but based on their analysis conclude that ``contamination is not a substantive confounder on the overall results." Given this disclosure of contamination, while it is impossible to determine the intersection between my dataset and their training or evaluation data, it is likely that some portions of my dataset could be present in their training data.

Nonetheless, I argue that the effect of any possible contamination is unobservable in my results. To investigate this, I looked at the accuracies when the gold answer was in its original position. As discussed in Section \ref{sec:experimental setup}, I prompt each question five times, rotating the answer choices so that the gold answer appears in each spot labeled (A)-(E) exactly once. As per Oren, Meister et al. \cite{oren2023proving}, if data contamination has a strong effect, we would expect that the model would perform better when the gold answer is in its original position (or what Oren, Meister et al. would call its ``canonical position"), i.e., the question is unaltered from its version on the web. 

So, for each question, I compute models' accuracy when the gold answer is in its canonical position and compare that to models' overall accuracy, averaged over all five answer positions. I calculate this for GPT-3.5's results over the whole dataset as well averaged over many several trials for each prompt setting. I also include this measure averaged over the numerous trials from our In-and-Out experiments, which will be later expanded on in Chapter \ref{ch:ino}. I also report the two accuracies over the one full run of Claude 2 and GPT-4 on my dataset. 

As we can see from the results in Table \ref{tab:contamination} below (displaying mean accuracies and standard deviations across many trials over many prompt settings), there is no significant bias toward the gold answer when it is in its canonical position as opposed to other positions in the rotation, with the canonical position in several trials even resulting in less accuracy than the rotated positions. This, along with the generally low accuracy of these models, gives me confidence that any potential downstream impact of dataset contamination on my evaluation is limited. 


\begin{table}[htbp]
  \caption{Model Accuracy in Canonical and Rotated Positions}   \label{tab:contamination}
    \renewcommand{\arraystretch}{1.2} 
    \vspace{-10pt} 
    \begin{center}
    \begin{tabular}{llcc}
    \hline \hline
    \toprule
    Model & Dataset & Canonical Position & Rotated Positions \\ \hline
    GPT-3.5 & In/Out Subset (32 trials) & $25.3 \pm 6.5$ & $23.4 \pm 4.0$ \\ \cline{2-4}
           & Full Dataset (5 Trials) & $21.6 \pm 2.6$ & $20.7 \pm 2.0$ \\ \hline
    \midrule
    GPT-4 & Full Dataset (1 Trial) & 32.2 & 33 \\\hline
    \midrule
    Claude 2 & Full Dataset (1 Trial) & 23.5 & 25.7 \\
    \bottomrule
    \end{tabular}%
    \end{center}
\end{table}

Furthermore, irrespective of contamination, analysis of the explanations generated by GPT-3.5 and GPT-4, as conducted in Section \ref{sec:explanation_analysis}, still provide valuable insights into the broad categories of errors in logic that the models are prone to.

\chapter{In-and-Out Problem Results and Analysis}\label{ch:ino}
Motivated by GPT-4's very strong accuracy-around 50\%—on In-and-Out questions, I dove deeper into this subset of the dataset. In this chapter, I summarize my efforts to push accuracy on this promising subset further, through a systematic exploring of Chain-of-Thought prompting and an implementation of Self-Reflection. Both achieve higher accuracy than zero-shot Chain-of-Thought prompting.

Additionally, in an effort to understand what \textit{kind} of logic GPT-3.5 and GPT-4 excel and fail at, I also qualitatively analyze a subset of this data in depth and discuss categories of logical errors GPT-3.5 and GPT-4 make on In-and-Out problems. Full implementation details are described in Section \ref{sec:ino_implementation}.

\section{Chain-of-Thought Trials}
\subsection{Chain-of-Thought improves GPT-3.5 Accuracy}
For each $k \in [5]$, I queried GPT-3.5 for 5-9 distinct prompts. Each prompt consisted of a different permutation of $k$ demonstrations picked from a pool of 5 demonstrations. Results are summarized in Table \ref{tab:cot-ino-summary} below, with full results for every prompt setting available in Appendix Table \ref{tab:cot-trials}. I report the accuracy of the best trial for each $k$ (in addition to the average accuracy) since I am interested in ascertaining the model's full capabilities given the right prompting. Out of the 32 total trials, four prompts resulted in over $28\%$ accuracy averaged over the In-and-Out Dataset, a notable improvement over 23.3\% for GPT-3.5 Zero-Shot Chain-of-Thought. This shows that Chain-of-Thought with examples targeted to the Logic Game Type can improve GPT-3.5 performance! 

Another curious result from Table \ref{tab:cot-ino-summary} is that while all values of $k$ had a prompt that resulted in a maximum accuracy of around $28\%$, the variance between prompts increases with $k$. Upon manual inspection, some multi-shot prompts fail because after seeing five demostrations, GPT-3.5 gets confused about what the user is asking it to do in response to the sixth question. Future work can further probe the higher variability of longer Chain-of-Thought prompts on logic games.

\begin{table}[htbp]
  \centering
  \caption{Summary of In-and-Out Chain-of-Thought Trials (GPT-3.5)} \label{tab:cot-ino-summary}
    \renewcommand{\arraystretch}{1.1} 
    \vspace{10pt} 
    \begin{tabular}{cccc}
    \toprule
    \hline \hline
    $k$ & Number of  & Average Accuracy  & Best Accuracy \\ 
    & Trials & (Over Trials) & (Best Trial)\\\hline \hline 
    \midrule
    0 (baseline) & 1 &   23.3 & 23.3 \\ \hline\hline
    1 & 5   &  26.4 & 28.8 \\ \hline
    2 & 5 & 25.3 & 28.8 \\ \hline
    3 & 9 & 23.7 & 26.5 \\ \hline
    4 & 5 & 20.0 & 28.4 \\ \hline
    5 & 9 & 21.1 & 28.4 \\ \hline\hline
    \bottomrule
    \end{tabular}%
\end{table}

From looking at the specific demonstrations associated with each prompt (enumerated in Table \ref{tab:cot-trials}), we can observe that the highest accuracy-achieving 3, 4, and 5-shot prompts all begin with the demonstration that achieved the best accuracy in the 1-shot setting. And in general, accuracy for a multi-shot prompt is correlated with the 1-shot accuracy of the first demonstration in that prompt, more than any other positions. If we consider the best-performing 1-shot demonstrations to be particularly ``relevant" context or examples for LLMs, this observation may be reflective of the ``Lost in the Middle" phenomenon \cite{liu2023lost}, wherein performance on a task degrades if ``relevant" information is in the middle as opposed to the beginning or end. Thus, this demonstrates the importance of carefully ordering prompt demonstrations for complex Chain-of-Thought settings such as this one, as ordering affects performance. 

However, unlike GPT-3.5, GPT-4 exhibited no improvement in multi-shot Chain-of-Thought. I evaluated GPT-4 on the dataset using four different Chain-of-Thought prompts (including some prompts that performed very well for GPT-3.5 and some prompts that did not, for diversity), and all resulted in greatly reduced accuracy, with the best prompt only achieving 38\% accuracy on the dataset  compared to 47\% in Zero-Shot Chain-of-Thought. This suggests a potential weakness of Chain-of-Thought prompting for the LSAT logic games task. I further probe this weakness in the remaining analysis in this section.

\subsection{Chain-of-Thought helps for easy questions but may hurt hard ones}
\label{sec:cot_weakness}
For each prompt, I looked at average accuracy by Game Type, by Problem Type, and by Problem Difficulty to see if there were any interesting trends or any relationship between the metadata of the prompt demonstrations and improved categories. One noteworthy trend I observed was that the effective (in terms of improving accuracy) Chain-of-Thought prompts very consistently increased accuracy greatly on ``Easy" questions of lower Problem Difficulties, but perhaps at the expense of increased variability or worse performance on ``Hard" questions. 

\begin{figure}
\caption{Multi-Shot Accuracy by Problem Difficulty} \label{fig:multishot_difficulty}
\vspace{0pt}
\centering
\includegraphics[width=\textwidth]{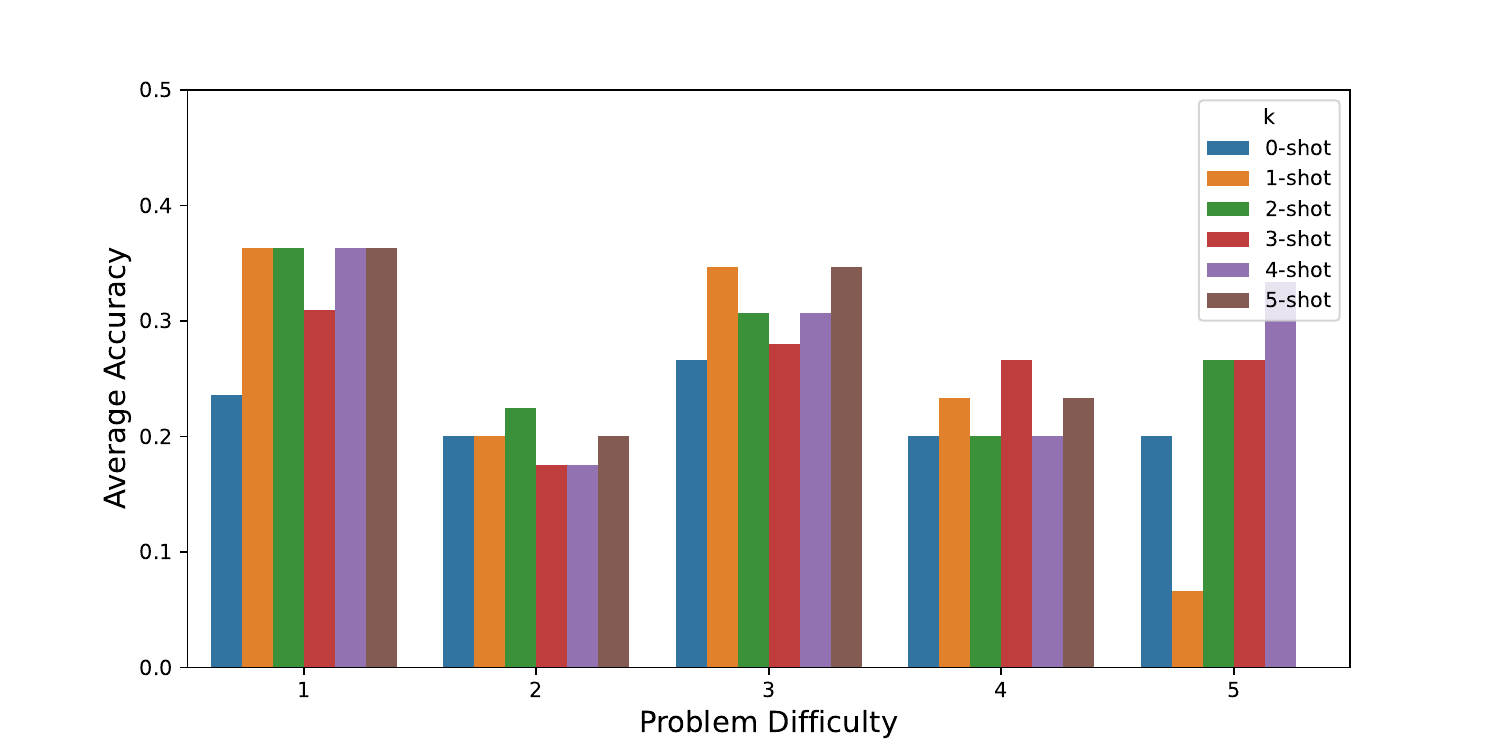}
\end{figure}

In Figure \ref{fig:multishot_difficulty}, I select the highest-accuracy multi-shot prompt for each $k$ and plot its average accuracy for each Problem Difficulty level. I also plot the zero-shot setting as a baseline. We see that all five settings improve accuracy considerably on Level 1 problems, from 23.6\% to up to 36.3\% for most prompts. For the most part, all difficulty levels improve, but we see that some prompts have a much lower accuracy on Level 5 than the zero-shot baseline. This trend is reflected also amongst prompts that were not effective in improving overall accuracy, with their weakness on Hard problems bringing down their accuracy.

One possible interpretation of this is that demonstrations can help models reason more thoroughly and avoid making silly mistakes on Easy questions. However, for hard questions, an overly structured demonstration may stifle the model's creativity in generating a reasoning path for the question. Especially given the particularly high diversity of reasoning paths that can be used across different LSAT logic games, the potentially stifling nature of a demonstration constitutes a real limitation of using Chain-of-Thought prompting for LSAT logic games \cite{zhong2023agieval}. It is difficult to craft a one-demonstration-fits-all example for multi-shot prompting, since problem types and problem solving routes look so different from problem to problem.

\subsection{Analysis of Positional Bias}
Positional bias toward particular answer choice labels has been documented for GPT-3.5 and GPT-4 in Multiple Choice Question answering \cite{pride}, where biases toward particular answer labels are supposedly dependent on the nature of the task. While I effectively mitigate this preference by prompting each question five times with the correct answer being assigned to each label once, the positional bias could still obscure the models' accuracy and capability, since I average accuracy over all answer answer labels. Thus, it is worth analyzing GPT-3.5's outputs to see if there is evidence of positional bias. I take the opportunity in this section to analyze positional bias, since in this set of experiments, we have many prompting outputs that we can aggregate over to see whether any bias is significant.

\begin{figure}
\caption{Average Accuracy For Each Answer Label} \label{fig:positional bias}
\includegraphics[width=\textwidth]{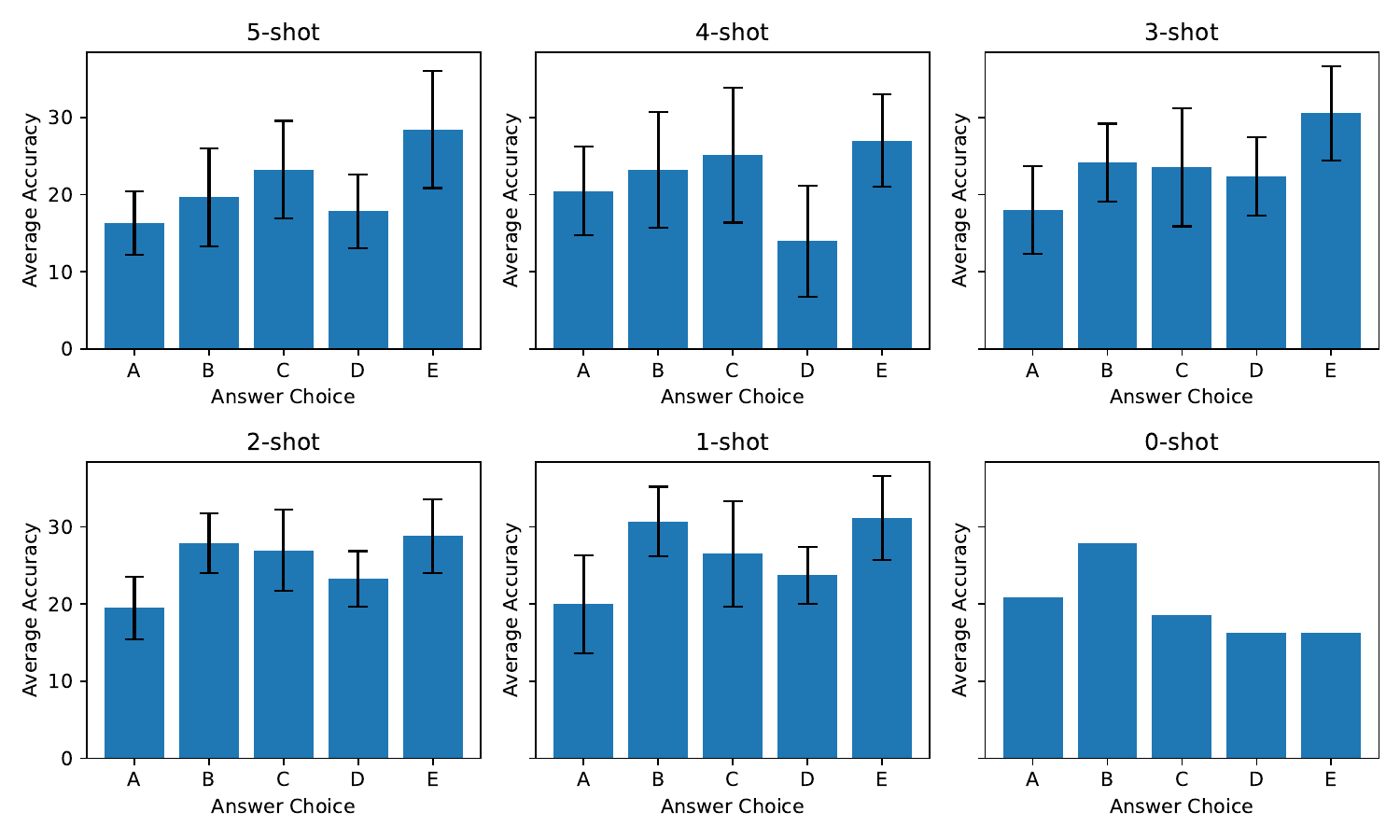}
\end{figure}

To measure positional bias I do the following: Since each answer label sees the correct answer once per question, for each answer label, we can compute the percentage of times that the model accurately judges the correct answer to be correct when it is in that label's position. In other words, we can say the accuracy for answer label ``(E)"'s is the percentage of times the model gets the question right when the correct answer choice is in position (E). If there is no positional bias, we would expect there to be no significant difference between the accuracies for each answer label. For each of the 32 trials, I compute the accuracy for each answer label. I then average these percentages over averaged over all the trials within each $k$-shot setting (indicated for each $k$ in Table \ref{tab:cot-ino-summary}). The results are displayed in Figure \ref{fig:positional bias}. 

We can see that for many of the multi-shot settings, there is a significant difference between the accuracy for answer label ``(E)" and the accuracy for answer label ``(A)", with answer label ``(E)" in some settings even reaching an average accuracy of 30\% as compared to option ``(A)" falling short at 17\%. Furthermore, the strength of the answer choice preference seems to also vary with the number of demonstrations, suggesting a possible relationship between the length of the prompt and the bias toward particular answer labels. This positional bias is a significant limitation on GPT-3.5's ability to answer Multiple Choice Questions.

\section{Self-Reflection}

For our best performing models, GPT-3.5 and GPT-4, I conducted the three-step Self-Reflection as shown in Figure \ref{fig:reflection}. For the first step, I take the each model's initial response to be its respective best-performing k-shot prompting setting on the In-and-Out subset. For GPT-3.5, responses were taken from a 1-shot setting, while for GPT-4, responses were taken from Zero-Shot Chain-of-Thought.
The results of the Self-Reflection experiments are summarized in Table \ref{tab:reflexion} below. ``Initial Accuracy" refers to the percentage of questions the model gets correct on the first try, ``Percent Recovered" refers to the percentage of questions the model gets correct after reflection and revision among those it initially got incorrect, and ``Final Accuracy" uses the ``Percent Recovered" to compute, among all problems, the percentage the model eventually gets right on the first or second try. As we can see, Self-Reflection improves performance considerably!

\begin{table}[htbp]
  \centering
  \caption{Self-Reflection Experiment Results (on 43-question Dataset)}
    \renewcommand{\arraystretch}{1.2} 
    \vspace{10pt} 
    \begin{tabular}{cccc}
    \toprule
    \hline \hline
    Model & Initial Accuracy & Percent Recovered & Final Accuracy \\ \hline \hline 
    \midrule
    GPT-3.5 & 21\% \scriptsize{ (9/43)}& 32\% \scriptsize{ (11/34)}& 47\% \scriptsize{ (20/43)} \\ \hline
    GPT-4 & 44\% \scriptsize{ (19/43)}& 46\% \scriptsize{ (11/24)}& 70\% \scriptsize{ (30/43)} \\
    \bottomrule
    \end{tabular}%
  \label{tab:reflexion}%
\end{table}
In obtaining these results, I hand-annotated initial answers to uncover the source of logical errors to be able to determine whether the reflections correctly pinpointed the source of error. For GPT-4, in all cases where the reflection correctly pinpointed the source of error, GPT-4 produced a correct revised answer (46\% of the time). However, for GPT-3.5, there were some cases where the model correctly pinpointed the source of the error, but had difficulty executing the correction. Despite only recovering the correct answer on 32\% of the questions it had initially gotten wrong, GPT-3.5 was able to produce a meaningful reflection (which I define as a specific reference to the incorrect logical turn) on its logical errors for 43\% of initially incorrect responses.

Additionally, qualitatively, I observe that GPT-4 often gives very specific reflections, pointing to particular answer choices or portions of its deductions where it went wrong as can be seen in Figure \ref{fig:reflection}, while GPT-3.5 often produces the same high-level, relatively vague reflection and plan, shown in Figure \ref{fig:reflection_gpt35}. Despite being vague, though, this reflection did occasionally result in a revised corrected answer.
\begin{figure}
\include{Figures/gpt3.5_reflection}
\caption{Common GPT-3.5 Reflection}\label{fig:reflection_gpt35}
\end{figure}
\begin{singlespace}
\end{singlespace}
Overall, though, the Final Accuracies in Table \ref{tab:reflexion} are very high! Of course, this improved Final Accuracy should be treated with cautious optimism, as the models did not get these questions correct on the first try and had to receive input from the environment that the first attempt was incorrect. Nonetheless, the high Final Accuracy and high Recovery Rate of GPT-3.5 and GPT-4 demonstrates an exciting capacity of both models to self-reflect, identify, and rectify highly complex logical errors!
\section{Analysis of Logical Errors in Explanations}
\label{sec:explanation_analysis}
In this section, I thoroughly analyze GPT-3.5 and GPT-4 explanations for a subset of 20 In-and-Out questions, looking at both the models' original outputs as well as corrected outputs after Self-Reflection. 

As helpful context, In-and-Out games involve assignments that place entities in an ``In" group or an ``Out" group (which take different names for each setup). They also involve many Implications as constraints. Since there are only two groups, these problems are easily mapped to Boolean logic, where each entity can be represented as a two-valued Boolean variable that can take on value 1 for ``In" or value 0 for ``Out" in an assignment (or vice versa). 

I break down the logical rules into formal logic \cite{logic} and categorize logical errors the models make, as well as note places where I believe the models excelled at deftly handling complicated logic. 

I synthesize my observations on errors and offer the following categorization of ten types of logical errors observed across both GPT-3.5 and GPT-4. I offer descriptions and supplement them with formal logic, where applicable. The first six relate to micro-level violations of basic rules about implications and deductions in formal logic, while the remaining errors are more high-level problem-solving errors:
\begin{enumerate}
    \item \textbf{Missing direct implication}. For an implication constraint of the form $A \implies B$ ($A$ implies $B$), errors that miss a direct implication produce a solution that clearly violate the direct implication. For example, for the constraint ``If Felipe is off stage, Jaclyn is also off stage," producing an assignment that has Felipe off stage but Jaclyn on stage would constitute such an error. Fortunately, such errors are rare for both GPT-3.5 and GPT-4, suggesting that these models are fairly adept at very basically applying forward direction implications.
    \item \textbf{Failure to apply the contrapositive}. This error involves only paying attention to the forward direction of an implication $A \implies B$, without applying the contrapositive $\neg B \implies \neg A$ (not $B$ implies not $A$). This error particularly surfaces in questions where models are asked to evaluate whether assignments of entities to the ``In" group in the In-and-Out problem is valid; the model may only pay attention to when the entities appear in the first predicate of the implication. For the same example constraint above, ``If Felipe is off stage [$A$], Jaclyn is also off stage [$B$]," such an error occurred when GPT-3.5 judged an assignment with Jaclyn on stage ($\neg B$) and Felipe off stage ($A$) to be valid, violating the contrapositive. GPT-3.5 makes such errors in three out of twenty evaluated examples, and GPT-4 makes these errors in two out of twenty examples, but there are also many solutions from both models that correctly apply the contrapositive.
    \item \textbf{Incorrect direction of implication}. Given an implication $A \implies B$, this error concludes $B \implies A$. The model's underlying error may be assuming bidirectionality of implication, or, simply relying on surface level similar text features without carefully paying attention to directionality. This error is surprisingly common— featuring four times for GPT-3.5 and three times for GPT-4. As an example, for the constraint ``4. If any of the women are on stage, Grant is also on stage," GPT-3.5 incorrectly deduces ``If Grant is on stage, then one of the women must also be on stage (from constraint 4)."
    \item \textbf{Mishandling logical inverses}. Given $A \implies B$, the models occasionally (twice for each model) incorrectly conclude that the inverse $\neg A \implies \neg B$ must be true. An example of GPT-4 making this error occurs in a problem with the constraint ``If Wong is on the team, both Myers and Yoder must also be." GPT-4 separately deduces that Wong cannot be on the team but then \textit{incorrectly} concludes that ``Myers cannot be on the team either because Wong's absence would prevent Myers from being included," indicating an occurrence of this inverse error.
    \item \textbf{Difficulty with Mutual Exclusivity (XOR)}. GPT-3.5 struggles with contraints involving mutual exclusivity (which can also be thought of as ``exclusive or's") appear, while GPT-4 handles them well (at least on the limited subset I analyze). For example, for the following constraint:\\ 
    \textit{Constraint:} ``Either Kayne or Novetzke, but not both, is assigned to one of the ambassadorships."\\
    \textit{GPT-3.5 Response:} ``(B) Venezuela: Novetzke, Yemen: Jaramillo, Zambia: Kayne. This choice satisfies all the constraints and is a valid assignment. We will keep this choice as a potential correct answer."\\
    This suggests a particular weakness in handling mutual exclusivity, as GPT-3.5 is typically good at verifying that answer choices satisfy direct implications (point 1 above). 
    \item \textbf{Ignoring the precondition}. In this error, which GPT-3.5 and GPT-4 each make once, given a constraint $A \implies B$, the model falsely assumes $A=1$ and concludes $B=1$. As an example, for question with the constraint, ``If Landon is assigned an ambassadorship, he must go to Zambia," without proving the precondition that Landon is assigned an ambassadorship, GPT-3.5 argues, ``As Landon must go to Zambia\ldots" and proceeds to make invalid deductions.
    \item \textbf{Incomplete Transitive Chain}. In this more high-level error, on questions that involve following a longer chain of implications transitively to make deductions, GPT-3.5 and GPT-4 may stop early. The reasons for stopping early sometimes involve aforementioned errors, like Failure to apply the contrapositive (point 2) or Missing direct implication (point 1).
    \item \textbf{Ignoring Cardinality Constraints}. In-and-Out question setups occasionally combine Implication constraints with Cardinality constraints on the size of the ``In" or ``Out" group. An example of this is a setup that begins, ``A corporate manager is selecting employees for a research team. The team will include at least four employees..." before listing Implication constraints. In all three questions I evaluate that require using a Cardinality constraint to solve the problem, GPT-3.5 and GPT-4 struggle to integrate the Cardinality constraint into their analysis of transitive implication deductions, and thus fail to reach proper conclusions.
    \item \textbf{Non Sequitur.} Both models make many (five for GPT-3.5, two for GPT-4) assertive logical conclusions that seem to come out of nowhere and do not logically follow from the constraint the model mentions to justify their conclusion. (See GPT-4's original response in Figure \ref{fig:reflection} for an example of GPT-4 making a non sequitur error.) Often in a non sequitur error, as seen in Figure \ref{fig:reflection}, when evaluating multiple answer choices, the model will refer to a constraint that involves the same entity as the answer choice and is seemingly relevant, but then make illogical conclusions based on that constraint.
    \item \textbf{Basic Inconsistencies}. This error refers to any time the model says something that directly contradicts what it said shortly before, within a short span of model output. For how basic these errors are, they are surprisingly \textit{extremely common}, occurring in six (out of twenty) responses for GPT-3.5 and three (out of twenty) for GPT-4. An example is when GPT-3.5 incorrectly deems an assignment in an answer choice possible as below, with the inconsistency italicized for emphasis:\\
    ``(C) \textit{Venezuela: Ong. Yemen: Kayne.} Zambia: Landon\ldots Constraint 3 is satisfied because even though Ong is assigned to Venezuela, \textit{Kayne is not assigned to Yemen}."
\end{enumerate}
Taken together, these errors highlight the types and variety of micro-level and macro-level logic that GPT-3.5 and GPT-4 still struggle with, especially when solving longer logic problems as in LSAT logic games.

\chapter{Conclusions and Future Work}\label{ch:conclusion}
In this thesis, I construct a novel dataset containing LSAT logic game questions and associated metadata that is rich for analysis. I explore and analyze the low accuracies achieved by Chain-of-Thought prompting \cite{wei2023chainofthought}, before applying a new framework called Reflexion \cite{reflexion}, which highlights the promising ability of LLMs to correct and revise their logical errors, with GPT-4 achieving 70\% accuracy and GPT-3.5 achieving 46\% accuracy on a subset of the data under this framework. I also draw attention to specific logical errors, some of them very basic, that GPT-3.5 and GPT-4 make in their response to ``In-and-Out" questions.

Future work may utilize my dataset as an evaluation tool for LLMs, or further analyze how LLM performance may differ based on Logic Game type and Problem Type. Future work may also apply Self-Reflection to the whole dataset to see if the improvement in performance generalizes to other Game Types. Finally, using the categories of logical errors I manually observe through human annotation as a base, future work may work on automating detection of these errors to quantify how often LLMs make these errors on a larger dataset.

This thesis represents a substantial in-depth analysis of the performance of LLMs on LSAT logic games and provides a solid starting ground for exciting future work.

\appendix
\chapter{Additional Sample Questions}\label{ch:sample Questions}
This appendix contains examples for a variety of Game Types, including variants on the three main types and hybrids. As earlier in the thesis, these examples are all from Khan Academy \cite{khan-academy}.

\begin{minipage}[t]{\textwidth}
    \vspace{0pt} 
    \begin{singlespace}
    \begin{promptbox}{Sequence With Conditional}
Each of exactly seven professors-Powell, Shihab, Taylor, Vaughan, Wood, Young, and Zabel-gives exactly one guest lecture in the literary theory course. The lectures are ordered from first through seventh, and their order must conform to the following:\\\\
-Powell lectures before Wood.\\
-Taylor lectures before Shihab.\\
-Vaughan lectures before Zabel.\\
-Shihab is no later than third.\\
-Young is not seventh.\\
-Powell lectures first if, but only if, Young lectures before Vaughan.\\\\
If Shihab lectures second and Zabel lectures fourth, then which one of the following could be true?\\
(A) Powell lectures sixth. (correct)\\
(B) Taylor lectures third.\\
(C) Vaughan lectures fifth.\\
(D) Wood lectures fifth.\\
(E) Young lectures third.
    \vspace{0.445cm}
    \end{promptbox}%
    \end{singlespace}
\end{minipage}%

\begin{minipage}[t]{\textwidth}
    \vspace{0pt} 
    \begin{singlespace}
    \begin{promptbox}{Grouping With Repetition}
In the Lifestyle, Metro, and Sports sections of tomorrow's newspaper, a total of six different photographs are to appear, exactly two photographs per section. Each of the available photographs was taken by one of three photographers: Fuentes, Gagnon, and Hue. Selection of the photographs is constrained by the following conditions:\\\\
-For each photographer, at least one but no more than three of that photographer's photographs must appear.\\
-At least one photograph in the Lifestyle section must be by a photographer who has at least one photograph in the Metro section.\\
-The number of Hue's photographs in the Lifestyle section must be the same as the number of Fuentes' photographs in the Sports section.\\
-None of Gagnon's photographs can be in the Sports section.\\\\
If one photograph in the Metro section is by Fuentes and one is by Hue, then which one of the following could be true?\\
(A) Both photographs in the Sports section are by Fuentes.\\
(B) Both photographs in the Lifestyle section are by Fuentes.\\
(C) Both photographs in the Lifestyle section are by Gagnon.\\
(D) One photograph in the Lifestyle section is by Gagnon and one is by Hue.\\
(E) Both photographs in the Lifestyle section are by Hue.
    \vspace{0.445cm}
    \end{promptbox}%
    \end{singlespace}
\end{minipage}%

\begin{minipage}[t]{\textwidth}
    \vspace{0pt} 
    \begin{singlespace}
    \begin{promptbox}{Sequence / Group (Hybrid)}
A new magazine is assigning photo essays to be featured in its first five monthly issues, one essay per issue. Three of the essays will have a rural theme and two will have an urban theme. Each essay will be assigned to a different one of five photographers: Fetter, Gonzalez, Howland, Jordt, and Kim. The assignment of photographers and themes to issues is subject to the following constraints:\\\\
-The essay featured in the first issue must have a rural theme.\\
-Kim's essay must be featured in the issue immediately preceding the issue in which Fetter's essay is featured.\\
-Fetter's essay cannot have the same type of theme as Kim's.\\
-Gonzalez's essay must be featured in the third issue.\\
-Jordt's essay must have an urban theme.\\\\
Which one of the following is an acceptable assignment of photographers to issues, listed in order from the first issue to the fifth?\\
(A) Fetter, Jordt, Gonzalez, Kim, Howland\\
(B) Gonzalez, Kim, Fetter, Jordt, Howland\\
(C) Howland, Kim, Gonzalez, Fetter, Jordt\\
(D) Jordt, Howland, Gonzalez, Kim, Fetter\\
(E) Kim, Fetter, Gonzalez, Jordt, Howland\\

    \vspace{0.445cm}
    \end{promptbox}%
    \end{singlespace}
\end{minipage}%

\begin{minipage}[t]{\textwidth}
    \vspace{0pt} 
    \begin{singlespace}
    \begin{promptbox}{In-and-Out / Sequence (Hybrid)}
In one week, Monday through Friday, a library's bookmobile will visit five of the following six neighborhoods: Hidden Hills, Lakeville, Nottingham, Oldtown, Park Plaza, and Sunnyside. Exactly one neighborhood will be visited on each of the five days, and none of the neighborhoods will be visited on more than one day. The bookmobile’s schedule must conform to the following conditions:\\\\
-Hidden Hills is visited, but not on Friday.\\
-If Oldtown is visited, then it is visited on the day immediately before Hidden Hills is visited.\\
-If Lakeville is visited, then it is visited on Wednesday.\\
-Nottingham and Sunnyside are both visited, but not on consecutive days.\\\\
If Hidden Hills is visited on Monday, which one of the following must be true? \\
(A) Lakeville is visited on Wednesday.\\
(B) Nottingham is visited on Tuesday.\\
(C) Park Plaza is visited on Thursday.\\
(D) Sunnyside is visited on Tuesday.\\
(E) Sunnyside is visited on Friday.

    \vspace{0.445cm}
    \end{promptbox}%
    \end{singlespace}
\end{minipage}%

\chapter{In-and-Out Chain-of-Thought Experiments}\label{ch:Ino Experiments}
To explore Chain-of-Thought for the In and Out dataset, for each $k \in [5]$, I queried GPT-3.5 for 5-9 distinct prompts, where each prompt consisted of a different permutation of $k$ demonstrations picked from a pool of 5 demonstrations. These demonstrations are ID-ed in the table by the index at which they appear in the Full Dataset. For reference, demo 248 consists of the In and Out sample question in Section \ref{sec:logic game type} and its explanation in Figure \ref{fig:explanation}. 

Table \ref{tab:cot-trials} below shows the average accuracy across the 43 questions in the In and Out dataset for each prompt setting, with the highest accuracy for each value of $k$ bolded. We can directly observe that all of the best-performing 3,4,5-shot prompts begin with the best-performing one-shot demonstrations (248 and 279). We can also see that there is considerably more variance in performance as the prompts get longer— validating the need to try many prompts before rushing to conclusions.

\begin{table}[h]
\caption{Accuracy of All In-and-Out Chain-of-Thought Trials}
\renewcommand{\arraystretch}{1.2} 
\vspace{10pt} 
\centering
\begin{tabular}{lll}
\hline\hline
\textbf{k} & \textbf{Demo IDs} & \textbf{Average} \\
& & \textbf{Accuracy (\%)}\\
\hline\hline
1 & 248 & \textbf{28.8} \\\cline{2-3}
  & 256 & 25.1 \\\cline{2-3}
  & 261 & 24.7 \\\cline{2-3}
  & 279 & 27.0 \\\cline{2-3}
  & 335 & 26.5 \\\cline{2-3}
\hline\hline
2 & 256, 248 & 23.3 \\\cline{2-3}
  & 261, 335 & 26.0 \\\cline{2-3}
  & 248, 279 & 24.7 \\\cline{2-3}
  & 279, 256 & \textbf{28.8} \\\cline{2-3}
  & 335, 261 & 23.7 \\\cline{2-3}
\hline\hline
3 & 335, 261, 248 & 26.0 \\\cline{2-3}
  & 248, 335, 261 & \textbf{26.5} \\\cline{2-3}
  & 256, 279, 335 & 20.5 \\\cline{2-3}
  & 335, 256, 279 & 24.2 \\\cline{2-3}
  & 279, 335, 256 & 23.7 \\\cline{2-3}
  & 261, 279, 256 & 22.8 \\\cline{2-3}
  & 256, 261, 279 & 21.4 \\\cline{2-3}
  & 279, 256, 261 & 24.7 \\\cline{2-3}
\hline\hline
4 & 256, 248, 279, 335 & 21.4 \\\cline{2-3}
  & 248, 335, 256, 261 & \textbf{28.4} \\\cline{2-3}
  & 261, 256, 279, 248 & 17.7 \\\cline{2-3}
  & 279, 261, 335, 256 & 26.0 \\\cline{2-3}
  & 256, 279, 248, 261 & 16.3 \\\cline{2-3}
\hline\hline
5 & 248, 335, 279, 256, 261 & \textbf{28.4} \\\cline{2-3}
  & 335, 279, 256, 261, 248 & 13.5 \\\cline{2-3}
  & 279, 256, 261, 248, 335 & 25.1 \\\cline{2-3}
  & 256, 261, 248, 335, 279 & 16.7 \\\cline{2-3}
  & 261, 248, 335, 279, 256 & 18.6 \\\cline{2-3}
  & 248, 335, 279, 256, 261 & 27.9 \\\cline{2-3}
  & 335, 279, 256, 261, 248 & 17.2 \\\cline{2-3}
  & 279, 256, 261, 248, 335 & 22.3 \\\cline{2-3}
  & 256, 261, 248, 335, 279 & 20.0 \\\cline{2-3}
\hline\hline
\end{tabular}
\label{tab:cot-trials}
\end{table}

\chapter{Code Statement}\label{ch:Code Statement}
In this appendix, I cite sources that I used as reference when coding. I referred to the GitHub repository for Reflexion \cite{reflexion} and adapted their prompts for my implementation of their framework, Self-Reflection. As a reference for writing code to query LLMS via the OpenAI API and the Anthropic API, I referred to the GitHub repository for QuRating \cite{wettig2024qurating} (but had to make many modifications due to the recent OpenAI API update). Finally, while GPT-3.5 may not be good at LSAT logic games, it is pretty good at writing short snippets of code for data processing and data analysis! I used it for help constructing routine functions for data pre-processing and for code to produce charts. Aside from these sources, I produced a substantial amount of code from scratch for this thesis.

\bibliographystyle{abbrv}
\bibliography{refs} \label{bib}

\end{document}